\documentclass[letterpaper, 10 pt, conference]{ieeeconf}

\usepackage{graphicx}
\usepackage{booktabs}
\usepackage[dvipsnames]{xcolor}
\usepackage{amsmath,amssymb,amsfonts}
\usepackage{hyperref}
\usepackage[capitalise]{cleveref}
\usepackage{multirow,multicol}
\usepackage{algorithmic}
\usepackage{textcomp}
\usepackage{xspace}
\usepackage{url}
\usepackage{dsfont}
\usepackage{balance}
\usepackage{soul}
\usepackage{todonotes}
\usepackage{pifont,bbding}
\usepackage{colortbl}
\usepackage{overpic}
\usepackage{nccmath}
\usepackage{courier}
\usepackage{array}
\usepackage{tabularray}
\usepackage{pifont}
\usepackage{microtype}
\usepackage{cite}

\usepackage{pgfplots}
\usepackage{pgfplotstable}

\usepackage{contour}
\usepackage{pgf,tikz}
\usepackage{tkz-euclide,tkz-graph}
\usetikzlibrary{calc}
\usepackage{pgfplots}
\pgfplotsset{
    compat=1.18,
    tick label style={font=\footnotesize},
    label style={font=\footnotesize},
    width=9cm,
    height=6cm}

\usepackage{subfiles}    
\usepackage{standalone}  
\usepackage{makecell}
\usepackage{caption}
\usepackage{comment}

\usepackage{tabularx}
\usepackage{array}
\IEEEoverridecommandlockouts
\title{\LARGE \bf
Free-form language-based robotic reasoning and grasping
}
\author{
Runyu Jiao$^{1,2,\dagger}$, 
Alice Fasoli$^{1,\dagger}$,
Francesco Giuliari$^{1}$,
Matteo Bortolon$^{1,2,3}$,
Sergio Povoli$^{1}$,\\
Guofeng Mei$^{1}$,
Yiming Wang$^{1}$,
Fabio Poiesi$^{1}$\\
$^{1}$Fondazione Bruno Kessler,
$^{2}$University of Trento,
$^{3}$Istituto Italiano di Tecnologia
}

\begin{document}

\newcommand{\cmark}{\ding{51}}
\newcommand{\xmark}{\ding{55}}
\newcommand{\warning}[1]{\textbf{\color{red!90}{#1}}}

\definecolor{myblue}{rgb}{0.0000,0.7490,1.0000}
\definecolor{myyellow}{rgb}{1.0000,0.7530,0.0000}
\definecolor{mygray}{rgb}{0.9000,0.9000,0.9000}
\definecolor{myazure}{rgb}{0.8509,0.8980,0.9412}

\newcommand{\fabiocomment}[1]{\todo[color=purple!20, inline, author=Fabio]{#1}}
\newcommand{\yimingcomment}[1]{\todo[color=red!20, inline, author=Yiming]{#1}}
\newcommand{\davidecomment}[1]{\todo[color=blue!20, inline, author=Davide]{#1}}
\newcommand{\francescocomment}[1]{\todo[color=green!20, inline, author=Francesco]{#1}}
\newcommand{\gmeicomment}[1]{\todo[color=yellow!20, inline, author=GMei]{#1}}
\newcommand{\sergiocomment}[1]{\todo[color=orange!20, inline, author=Sergio]{#1}}
\newcommand{\alicecomment}[1]{\todo[color=pink!20, inline, author=Alice]{#1}}
\newcommand{\runyucomment}[1]{\todo[color=gray!20, inline, author=Runyu]{#1}}

\newcommand{\gmei}[1]{\textcolor{red}{GMEI: #1}}

\newcommand{\acronym}{tev00\xspace}
\newcommand{\vlm}{VLM\xspace}
\newcommand{\vlms}{VLMs\xspace}

\def\methodshort{OURSHORT\xspace}
\def\method{OursLongName\xspace}

% Custom commands
\makeatletter
\DeclareRobustCommand\onedot{\futurelet\@let@token\@onedot}
\def\@onedot{\ifx\@let@token.\else.\null\fi\xspace}

\def\eg{\emph{e.g}\onedot} \def\Eg{\emph{E.g}\onedot}
\def\ie{\emph{i.e}\onedot} \def\Ie{\emph{I.e}\onedot}
\def\cf{\emph{c.f}\onedot} \def\Cf{\emph{C.f}\onedot}
\def\etc{\emph{etc}\onedot} \def\vs{\emph{vs}\onedot}
\def\wrt{w.r.t\onedot} \def\dof{d.o.f\onedot}
\def\etal{\emph{et al}\onedot}
\makeatother

% Model names
\newcommand{\gpt}{GPT-4o\xspace}
\newcommand{\molmo}{Molmo\xspace}

% Method names
\def\ourmethodshort{\textcolor{darkgray}{FreeGrasp}\xspace}
\def\ourmethod{\textcolor{orange}{OursLongName}\xspace}
\def\gpt{GPT-4o\xspace}

% Dataset names
\def\ourdatasetshort{\textcolor{darkgray}{FreeGraspData}\xspace}
\def\ourdataset{free-from language grasping dataset\xspace}

\twocolumn[{%
    \renewcommand\twocolumn[1][]{#1}%
    \maketitle
    \thispagestyle{empty}
    \vspace{-8mm}
    \begin{center}
\captionsetup{type=figure}
\includegraphics[width=1.0\linewidth]{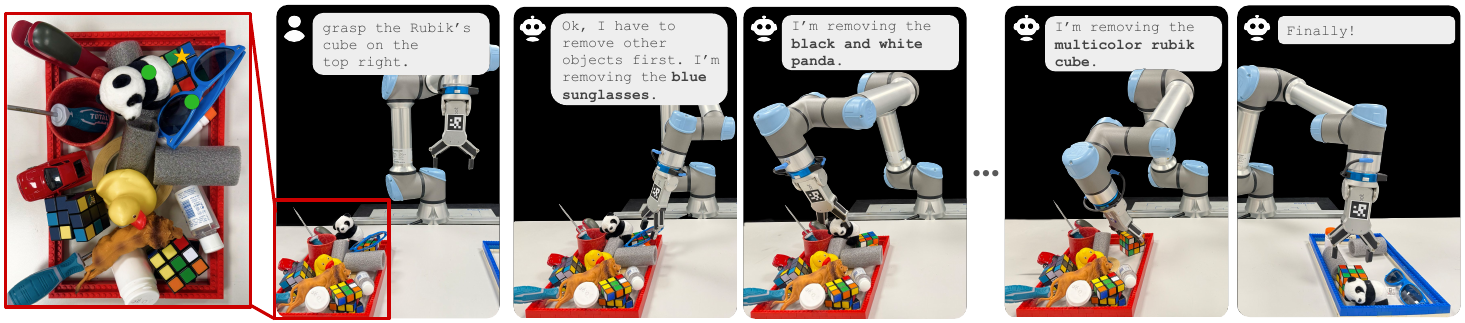}
\caption{To enable a human to command a robot using free-form language instructions, our method leverages the world knowledge of Vision-Language Models to interpret instructions and reason about object spatial relationships.
This is important when the target object ({\color{yellow}\FiveStar}) is not directly graspable, requiring the robot to first identify and remove obstructing objects ({\tiny{\color{green}\CircleSolid}}).
By optimizing the sequence of actions, our approach ensures efficient task completion.
}

\label{fig:teaser}
\end{center}
}]

\pagestyle{empty}

\let\thefootnote\relax\footnotetext{$^\dagger$Equal contribution.
This project was supported by Fondazione VRT under the project Make Grasping Easy, PNRR ICSC National Research Centre for HPC, Big Data and Quantum Computing (CN00000013), and
FAIR - Future AI Research (PE00000013), funded by NextGeneration EU.}

\begin{abstract}

Performing robotic grasping from a cluttered bin based on human instructions is a challenging task, as it requires understanding both the nuances of free-form language and the spatial relationships between objects.
Vision-Language Models (VLMs) trained on web-scale data, such as GPT-4o, have demonstrated remarkable reasoning capabilities across both text and images.
\textit{But can they truly be used for this task in a zero-shot setting? And what are their limitations?}
In this paper, we explore these research questions via the free-form language-based robotic grasping task and propose a novel method, FreeGrasp, leveraging the pre-trained VLMs’ world knowledge to reason about human instructions and object spatial arrangements.
Our method detects all objects as keypoints and uses these keypoints to annotate marks on images, aiming to facilitate GPT-4o’s zero-shot spatial reasoning.
This allows our method to determine whether a requested object is directly graspable or if other objects must be grasped and removed first.
Since no existing dataset is specifically designed for this task, we introduce a synthetic dataset FreeGraspData by extending the MetaGraspNetV2 dataset with human-annotated instructions and ground-truth grasping sequences.
We conduct extensive analyses with both FreeGraspData and real-world validation with a gripper-equipped robotic arm, demonstrating state-of-the-art performance in grasp reasoning and execution. Project website: \href{https://tev-fbk.github.io/FreeGrasp/}{https://tev-fbk.github.io/FreeGrasp/}.

\end{abstract}

%%%%%%%%%%%%%%%%%%%%%%%%%%%%%%%%%%%%%%%%%%%%%%%%%%%%%%%%%%%%%%%
%%%%%%%%%%%%%%%%%%%%%%%%%%%%%%%%%%%%%%%%%%%%%%%%%%%%%%%%%%%%%%%
%%%%%%%%%%%%%%%%%%%%%%%%%%%%%%%%%%%%%%%%%%%%%%%%%%%%%%%%%%%%%%%
\section{Introduction}\label{sec:intro}

Vision-Language Models (\vlms) encode vast semantic knowledge about the world, which is essential for interpreting nuanced human free-form language instructions \cite{openai2024gpt4technicalreport}.
This valuable source of information can enable robots to operate effectively in diverse, unseen, and unstructured environments \cite{Ahn2022,Chen2023}.
Interpreting human instructions is only one part of the challenge; grounding the retrieved information in the real world is equally critical. 
This process requires spatial reasoning~\cite{jointmodeling2023,SpaTiaL2023}, \ie~understanding and acting upon spatial relationships between objects \cite{SpatialVLM2024,D3GD2025}.
By achieving these capabilities together, we can tackle novel problems and enable embodied agents to perform object manipulation and rearrangement in less-controlled environments \cite{multiteacher2024}.
% \cite{multiteacher2024,plot2024,twostagegrasping2023,shiftingobjects2019}.

Recently, several studies have explored the use of pre-trained large-scale models for robotic control \cite{ReKep2024, moka2024, qian2024thinkgrasp}.
Building on this direction, we investigate how to leverage \vlms' knowledge to enable a robot to interpret human commands.
However, unlike previous works, our focus is specifically on evaluating the robustness of \vlms to free-form language instructions combined with spatial reasoning capabilities for robotic grasping.
Our setup, illustrated in \cref{fig:teaser}, consists of a robotic arm with a two-finger gripper, an exocentric RGB-D camera, a red bin containing various objects in clutter, and an empty blue bin where the grasped objects are put.
The objects in the red bin are arranged haphazardly, and multiple instances of the same object may be present, e.g.,~Rubik's cubes in this case.
Our robotic system is commanded by users through instructions, \eg \texttt{grasp the Rubik's cube on the top right}, with the goal of identifying which objects must be removed first in order to grasp the user-specified target.
The system should minimize the number of grasping actions to complete the task efficiently.

To this end, we introduce a free-form language-based robotic grasping approach, \textit{\ourmethodshort}, which leverages \vlms' world knowledge for reasoning about human instructions and object spatial arrangement.
Specifically, all visible objects within a container (or bin) are detected as keypoints, and the input image is augmented using mark-based visual prompting \cite{moka2024}.
The augmented image is then analyzed by \gpt \cite{openai2024gpt4technicalreport}, which interprets the free-form language instruction and reasons spatially about which object to interact with.
GPT-4o’s output is used to segment the input image and identify the correct object instance, particularly when multiple instances of the same object class are present.
Depth information and the segmented object instance are then used to estimate the grasp pose \cite{fang2020graspnet}.
Note that, our study aims to \textit{investigate} the reasoning capabilities of pre-trained models, hence, we do not train any model of our pipeline on task-specific data.
We validate our approach through both synthetic data and real-world robotic experiments.
Since no specific dataset exists to evaluate this task setup, we extended an existing bin-picking dataset, MetaGraspNetV2 \cite{MetaGraspNetV22024}, creating a new evaluation dataset named \ourdatasetshort.
Specifically, we involved ten human participants to annotate free-form language instructions.
\ourdatasetshort consists of scenarios with different task difficulties (Easy, Medium, Hard) based on obstruction conditions, each contains cases with and without object ambiguity, \ie the presence of multiple instances of the same class.
In total, \ourdatasetshort features 300 scenarios, each is associated with three annotated instructions from different annotators.
We compare the performance of \ourmethodshort against the recent approach ThinkGrasp~\cite{qian2024thinkgrasp}, which also features grasp reasoning with \gpt.
Experiments show that \ourmethodshort can robustly interpret human instructions and infer the actions to grasp a requested object more effectively than ThinkGrasp.
In summary, our contributions are:
\begin{itemize} 
    \item We introduce a novel robotic grasping setup where reasoning is key to interpreting human free-form language instructions and understanding spatial arrangement. 
    \item We propose a novel method that leverages \vlms' world knowledge to address this task setup without additional training on task-specific data. 
    \item We construct a new evaluation dataset that extends MetaGraspNetV2 \cite{MetaGraspNetV22024} to validate the effectiveness of \ourmethodshort in this novel setup. 
    \item Our robotic experiments confirm the advantages of \ourmethodshort in reasoning and grasping under real-world conditions with clutter and object ambiguity.
\end{itemize}

\section{Related works}\label{sec:related}
\noindent\textit{\vlm-driven robotic grasping.}
Multimodal learning has significantly enhanced robots' ability to interpret language instructions and perform grasp reasoning tasks. Research has focused on task-oriented grasping and spatial affordance prediction through \vlms. For example, RoboPoint~\cite{yuan2024robopoint} integrates \vlms to predict spatial affordances, improving object interaction understanding. SpatialVLA~\cite{qu2025spatialvla} enhances Vision-Language-Action (VLA) models with spatial representations, improving task planning and execution. GLOVER~\cite{ma2024glover} advances open-vocabulary reasoning for task-oriented grasping, enabling robots to perform diverse tasks without task-specific training.
To improve spatial reasoning, SpatialCoT~\cite{liu2025spatialcot} combines coordinate alignment with chain-of-thought reasoning for improved embodied task planning. 
SpatialPIN~\cite{spatialPIN2024} refines this by integrating 3D priors and advanced prompting techniques, enhancing object recognition, scene understanding, and navigation.
% %
% Authors in \cite{wang2024can,lu2023vl} focus on enabling robots to follow natural language instructions for manipulation tasks. 
ThinkGrasp~\cite{qian2024thinkgrasp}, somewhat related to our work, combines visual recognition and language-based reasoning to improve part grasping in cluttered environments.
Unlike ThinkGrasp, our method, \ourmethodshort, can handle free-form language instructions perform spatial reasoning more effectively.
Moreover, \ourmethodshort can interpret complex language descriptions to resolve object ambiguities, ensuring accurate identification of target objects even in cluttered or ambiguous environments.

\vspace{1mm}
%-------------------------------------------------
\noindent\textit{Learning-based and traditional robotic grasping.}
Traditional methods, which often rely on heuristic or geometric approaches, may struggle in complex scenarios.
To address these limitations, researchers have increasingly emphasized relationship reasoning to enhance robotic systems' robustness and adaptability, enabling them to better understand object interactions and perform tasks in cluttered environments.
A key milestone is VMRN~\cite{zhang2018visual}, which introduced a framework and dataset for predicting object manipulation relationships, highlighting the role of relationship reasoning in robotics. 
HSRN~\cite{wu2023prioritized} advanced this by proposing a hierarchical stacking relationship prediction method, optimizing manipulation in cluttered, occluded settings.
Large-scale datasets like REGRAD~\cite{zhang2022regrad} and MetaGraspNetV2~\cite{MetaGraspNetV22024} have been pivotal. REGRAD~\cite{zhang2022regrad} focuses on safe, object-specific grasping in clutter, while MetaGraspNetV2~\cite{MetaGraspNetV22024} offers a comprehensive dataset for bin picking, integrating relationship reasoning and dexterous grasping. 
These datasets have enabled more sophisticated grasp planning strategies, improving performance in complex scenarios.
Recent works~\cite{rabino2025modern,tang2022relationship} integrate visual relationship reasoning and semantic scene understanding into grasp planning. 
D3GD~\cite{rabino2025modern} proposes a scalable method for real-time object relationship reasoning. RelationGrasp~\cite{liu2024relationgrasp} introduces prompt learning for grasp detection and relationship prediction, showing promise in unstructured environments. 
GOAL~\cite{li2024grasping} addresses occlusion by combining occlusion-aware perception with relationship reasoning, significantly improving grasp success in highly cluttered scenes.

\section{Our method}\label{sec:method}
%-------------------------------------------------
\begin{figure*}[t]
\centering
\includegraphics[width=\textwidth]{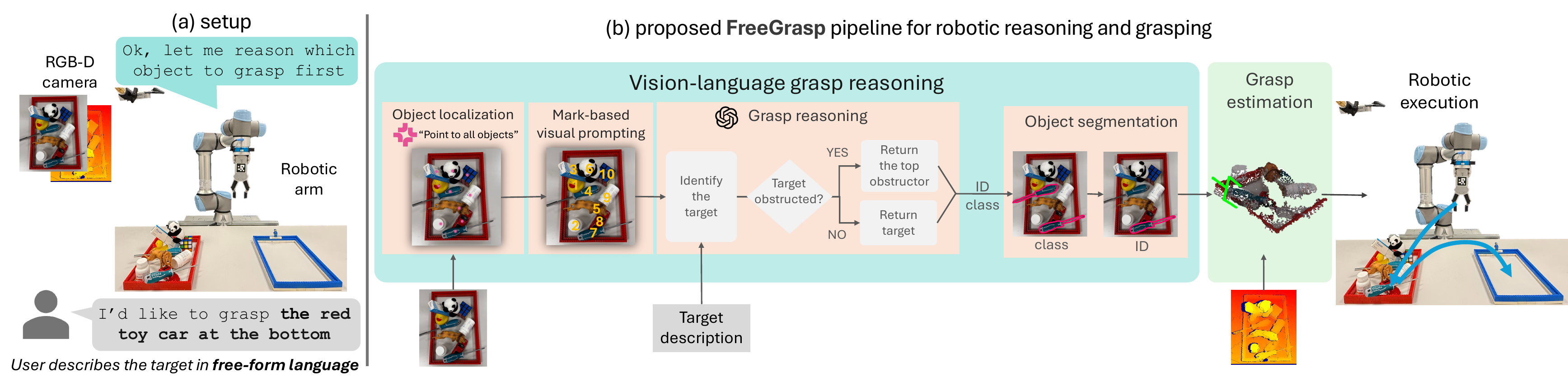}
\vspace{-5mm}
\caption{\ourmethodshort pipeline. (a) The setup considered for the robotic reasoning and grasping task and (b) the proposed pipeline that leverages pre-trained \vlms in a zero-shot manner without additional training. }
\label{fig:method}
\vspace{-4mm}
\end{figure*}
%-------------------------------------------------

%%%%%%%%%%%%%%%%%%%%%%%%%%%%%%%%%%%%%%%%%%%%%%%%%%%%%%%%%%%%%%
%%%%%%%%%%%%%%%%%%%%%%%%%%%%%%%%%%%%%%%%%%%%%%%%%%%%%%%%%%%%%%
\subsection{Overview}

We address the robotic reasoning and grasping task and propose an effective and modular pipeline by leveraging the world knowledge of pre-trained \vlms without additional training. 
Our setup consists of a bin containing objects organized haphazardly, an exocentric RGB-D camera with top-down view, a robotic manipulator, and a user who provides a free-form language instruction to request the robot to grasp a target object, e.g.~``\texttt{grasp the red toy car at the bottom}.''
\cref{fig:method} shows the diagram of \ourmethodshort.

An episode starts upon receiving the user's instruction. 
At each step, the pipeline starts with the proposed \textit{vision-language grasp reasoning} module.
With the RGB observation, the module first localizes all the objects within the container, forming a holistic understanding of the scene \cite{molmo2024}. 
To facilitate VLMs in visual spatial reasoning, we augment the visual prompt by annotating identity (ID) marks for each localized object on the input image.
Then, we feed the mark-based visual prompt to \gpt~\cite{openai2024gpt4technicalreport} to reason about object spatial relationships. 
\gpt determines whether the target object is free of obstruction, thus directly graspable, or is obstructed by other objects that must be removed first.
\gpt's output consists of the ID and class name of the next object to be grasped, along with information indicating whether it is the target object.  
With the class name and ID, the object segmentation model further segments the output object~\cite{langsam2024}.
The \textit{grasp estimation} module determines the most suitable grasp pose for picking the segmented object~\cite{fang2020graspnet}.
Lastly, the robotic arm plans and executes its trajectory to grasp the object and place it in a predefined location, \eg, another bin.
After each step, the positions of objects may change. The pipeline thus repeats with a new RGB-D acquisition. 
The episode for the robotic reasoning and grasping task terminates until the user-specified target object is being grasped or certain criteria is met.

%%%%%%%%%%%%%%%%%%%%%%%%%%%%%%%%%%%%%%%%%%%%%%%%%%%%%%%%%%%%%%
%%%%%%%%%%%%%%%%%%%%%%%%%%%%%%%%%%%%%%%%%%%%%%%%%%%%%%%%%%%%%%
\subsection{Vision-language grasp reasoning}
Given the RGB image capturing the scene and the user's free-form language instruction, this module leverages state-of-the-art \vlms to visually ground objects and reason about their relationship to determine the object to grasp, without additional model training.

\begin{figure}[t]
\centering
\includegraphics[width=\columnwidth]{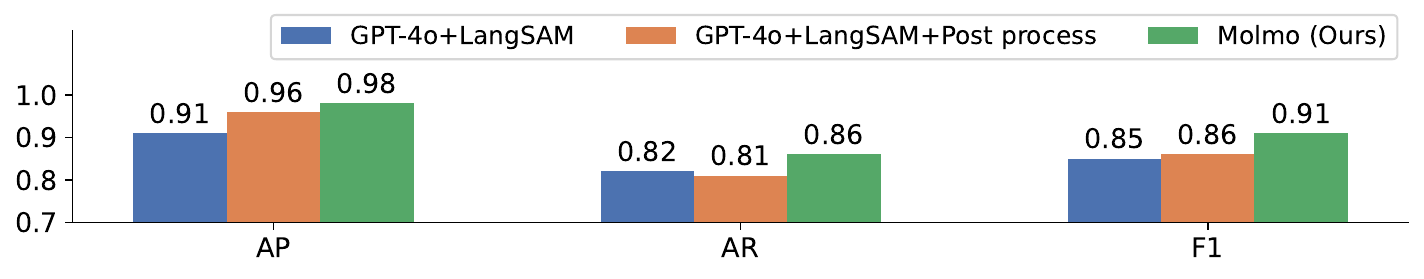}
\vspace{-5mm}
\caption{Object localization performance with different \vlm-based method on MetaGrasNetv2.}
\label{tab:detection}
\vspace{-4mm}
\end{figure}

\vspace{1mm}
%%%%%%%%%%%%%%%%%%%%%%%%%%%%%%%%%%%%%%%%%%%%%%%%%%%%%%%%%%%%%%
\noindent\textit{Object localization.}
There exist several \vlm-based open-vocabulary object localization methods like \cite{gdino2024}.
We mainly explored two options, and empirically selected the best-performing one for our task setup.
One option is to first prompt a \vlm (\eg~\gpt \cite{openai2024gpt4technicalreport}) for obtaining a list of object names in the bin and then use open-vocabulary segmentation models~(\eg LangSAM \cite{langsam2024}) for object localization, similar to~\cite{mei2025vocabulary}.
Alternatively, we can directly prompt a \vlm with visual grounding capability (\eg Molmo~\cite{molmo2024}) to point at all objects in the bin.
The output is a list of 2D coordinates, which are typically around the centers of the corresponding objects (\cref{fig:method}).
We evaluated the object localization performance with MetaGraspNetv2~\cite{MetaGraspNetV22024} and report the Average Precision (AP), Average Recall (AR), and F1 score in \cref{tab:detection}.
Directly leveraging Molmo leads to the better localization performance than the alternative with \gpt and LangSAM. 
The latter is prone to produce duplicate or fragmented masks among nearby objects, which is non-trivial to address with simple post-processing, \eg removing excessively small or large masks. 
Therefore, we selected Molmo for pointing.

%-------------------------------------------------
%-------------------------------------------------

\begin{figure*}[t]
\centering
\begin{overpic}[width=0.95\textwidth]{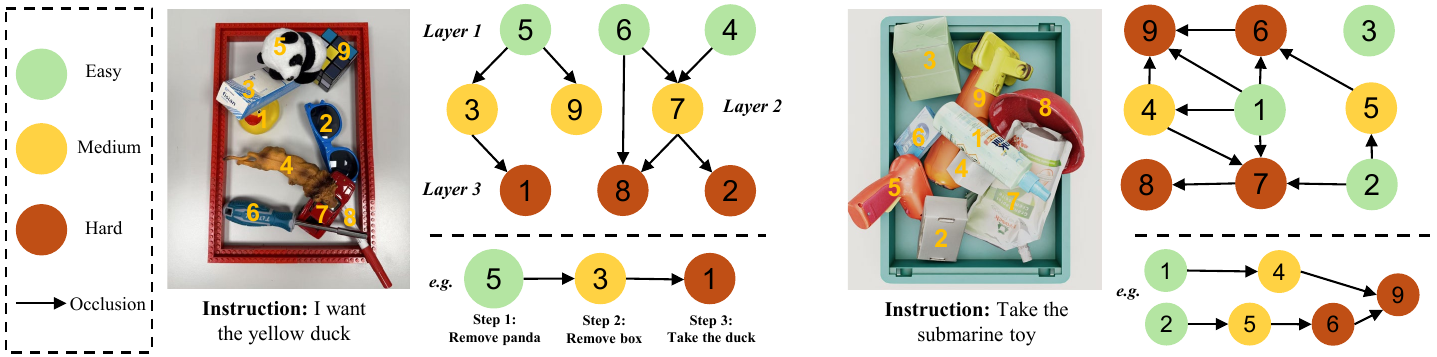}
    \put(25,-1.5){\small (a) Real world}
    \put(72,-1.5){\small (b) FreeGraspData}
\end{overpic}
\vspace{2mm}
\caption{Occlusion graphs in representative scenarios.
(a) Real-world scene and its corresponding occlusion graph.
(b) Synthetic scene from FreeGraspData and its occlusion graph.
Objects are colored by difficulty, and directed edges indicate occlusion dependencies.
Example task instructions and step-by-step action plans are shown beneath each graph.}
\label{fig:graph}
\vspace{-4mm}
\end{figure*}

\vspace{1mm}
%%%%%%%%%%%%%%%%%%%%%%%%%%%%%%%%%%%%%%%%%%%%%%%%%%%%%%%%%%%%%%
\noindent\textit{Mark-based visual prompting.}
\vlms can reason more effectively when inputs are presented in a multiple-choice format~\cite{somyang2023,moka2024}.
Hence, we assign a unique number to each object, and augment the input RGB image by annotating the numbered markers at the 2D object coordinates provided by the previous step (\cref{fig:method}). 
Then, we input the mark-based visual prompt into the reasoning \vlm for grasp reasoning.

\vspace{1mm}

%%%%%%%%%%%%%%%%%%%%%%%%%%%%%%%%%%%%%%%%%%%%%%%%%%%%%%%%%%%%%%
\noindent\textit{Grasp reasoning.} 
The \vlm must determine which object to grasp in each episode, which is the most essential yet challenging aspect of our proposed pipeline.
Effective grasp reasoning requires the \vlm to understand both:
i) free-form language instructions and
ii) spatial relationships, particularly in terms of obstructions.
We leverage \gpt, as it demonstrated the strongest reasoning capabilities among the \vlms we tested.

Free-form language instructions from different users can vary dramatically when referring to the same object, due to different life backgrounds of users.
For example, in our experiments, the target object in \cref{fig:dataset} (bottom left) is referred by different users as \texttt{juice box}, \texttt{juice}, or \texttt{refill pouch}.
Such ambiguous (even wrong) instructions pose significant challenges to the \vlm to reason on the correct target object. To resolve ambiguities, the capability in extracting relevant contextual information is critical, including spatial references (\eg \texttt{top left corner}) or part names (\eg \texttt{pouch}).
In the case of user instruction \texttt{the juice box on the top left corner}, \gpt responds as \texttt{I don't see a juice box in the top left corner. If you mean the object labeled "3", it is free of obstacles}.

For obstruction reasoning, the \vlm needs to understand spatial relationships among the target object and its nearby objects. This is particularly challenging to infer from a single image when objects are layered on top of each other. Despite \gpt being the best \vlm in terms of reasoning, its performance in this regard remains fairly poor. 
To improve occlusion reasoning, we initially explored recent advances in Chain-of-Thought (CoT) reasoning~\cite{cot2022} by first generating a detailed description of all objects and then summarizing it into a scene graph~\cite{3dgrand2024} that captures object adjacency.
Surprisingly, this approach did not enhance performance.
We observed that the model frequently hallucinated spatial relationships and produced inconsistent solutions for the same prompt, even when the \vlm's temperature was set to zero.
Instead, we found that an effective strategy is to \textit{contextualize the task} within the prompt by explicitly defining key aspects:
the setup (\eg a robotic arm with a parallel gripper for bin picking),
the objective (\eg grasp the target without obstruction),
the reasoning logic (\eg determine actions based on obstructions),
and the possible actions (\eg return the target or identify the top obstructor).
With this structured prompt, the \vlm reliably returns the ID and class name of the object to grasp.

\vspace{1mm}

%%%%%%%%%%%%%%%%%%%%%%%%%%%%%%%%%%%%%%%%%%%%%%%%%%%%%%%%%%%%%%
\noindent\textit{Object segmentation.} We segment the object instance identified in the image. 
For instance segmentation, simply using the 2D coordinate given by the object ID as visual prompt to class-agnostic segmentation models, \eg SAM~\cite{sam2023}, results in inaccurate instance masks with little semantic control. 
We thus first perform semantic segmentation using LangSAM~\cite{langsam2024} (\cref{fig:method}) with the class name. 
As there might be multiple instances with the same class, we then use the object ID to filter the instance mask of interest.

%%%%%%%%%%%%%%%%%%%%%%%%%%%%%%%%%%%%%%%%%%%%%%%%%%%%%%%%%%%%%%
%%%%%%%%%%%%%%%%%%%%%%%%%%%%%%%%%%%%%%%%%%%%%%%%%%%%%%%%%%%%%%
\subsection{Grasp estimation}
Lastly, we estimate the object's grasp pose with respect to the robotic arm. 
Once the object instance is segmented, we use GraspNet~\cite{fang2020graspnet} to regress the grasp pose. 
GraspNet operates on a colored point cloud, therefore we first lift the RGB and depth image into 3D using the camera's intrinsic parameters. 
Then, through the instance mask, we crop the point cloud to retain only the region corresponding to the object of interest and select the grasp pose estimated by GraspNet with the highest confidence.

% \begin{table*}[t]
% \centering
% \renewcommand{\arraystretch}{0.9}
% \tabcolsep 8.5pt
% \caption{Experiments on \ourdatasetshort. Higher metric values (SSR and RSR) indicate better performance. Best performance under each setting is in \textit{italic}.}
% \vspace{-2mm}
% \label{tab:results_synt}
% % \resizebox{\linewidth}{!}
% {%
% \begin{tabular}{lccccccccc}
% \toprule
% \multirow{2}{*}{Method} & \multirow{2}{*}{Reas.} & \multirow{2}{*}{Segm.}& \multirow{2}{*}{Metric} & \multicolumn{2}{c}{Easy} & \multicolumn{2}{c}{Medium} & \multicolumn{2}{c}{Hard} \\
% \cmidrule(lr){5-6} \cmidrule(lr){7-8} \cmidrule(lr){9-10}
%  & & & & w/o Amb. & w Amb. & w/o Amb. & w Amb. & w/o Amb. & w Amb. \\
% \midrule
% ThinkGrasp~\cite{qian2024thinkgrasp} &\checkmark & \checkmark &SSR & 0.63±0.02 & 0.46±0.02 & 0.13±0.03 & 0.16±0.02 & 0.05±0.02 & \textit{0.15±0.02} \\
% \ourmethodshort  & \checkmark & \checkmark &SSR & \textit{0.64±0.03} & \textit{0.64±0.04} & \textit{0.40±0.04} & \textit{0.35±0.02} & \textit{0.13±0.01} & 0.13±0.02 \\
% \midrule
% \ourmethodshort & \checkmark(GT) & & RSR& \textit{0.83±0.02} & 0.77±0.02 &\textit{0.46±0.03} & 0.31±0.06 & 0.21±0.01 & \textit{0.16±0.04} \\
% \ourmethodshort & \checkmark (Molmo) & & RSR & 0.83±0.06 & \textit{0.85±0.07} & 0.46±0.04 & \textit{0.33±0.04} & \textit{0.22±0.04} & 0.15±0.04 \\
% \bottomrule
% \end{tabular}
% }
% \vspace{-4mm}
% \end{table*}

%%%%%%%%%%%%%%%%%%%%%%%%%%%%%%%%%%%%%%%%%%%%%%%%%%%%%%%%%%%%%%
%%%%%%%%%%%%%%%%%%%%%%%%%%%%%%%%%%%%%%%%%%%%%%%%%%%%%%%%%%%%%%
%%%%%%%%%%%%%%%%%%%%%%%%%%%%%%%%%%%%%%%%%%%%%%%%%%%%%%%%%%%%%%
\section{Free-form language grasping dataset}\label{sec:dataset}

\begin{figure}[t]
    \centering
    \includegraphics[width=0.45\textwidth]{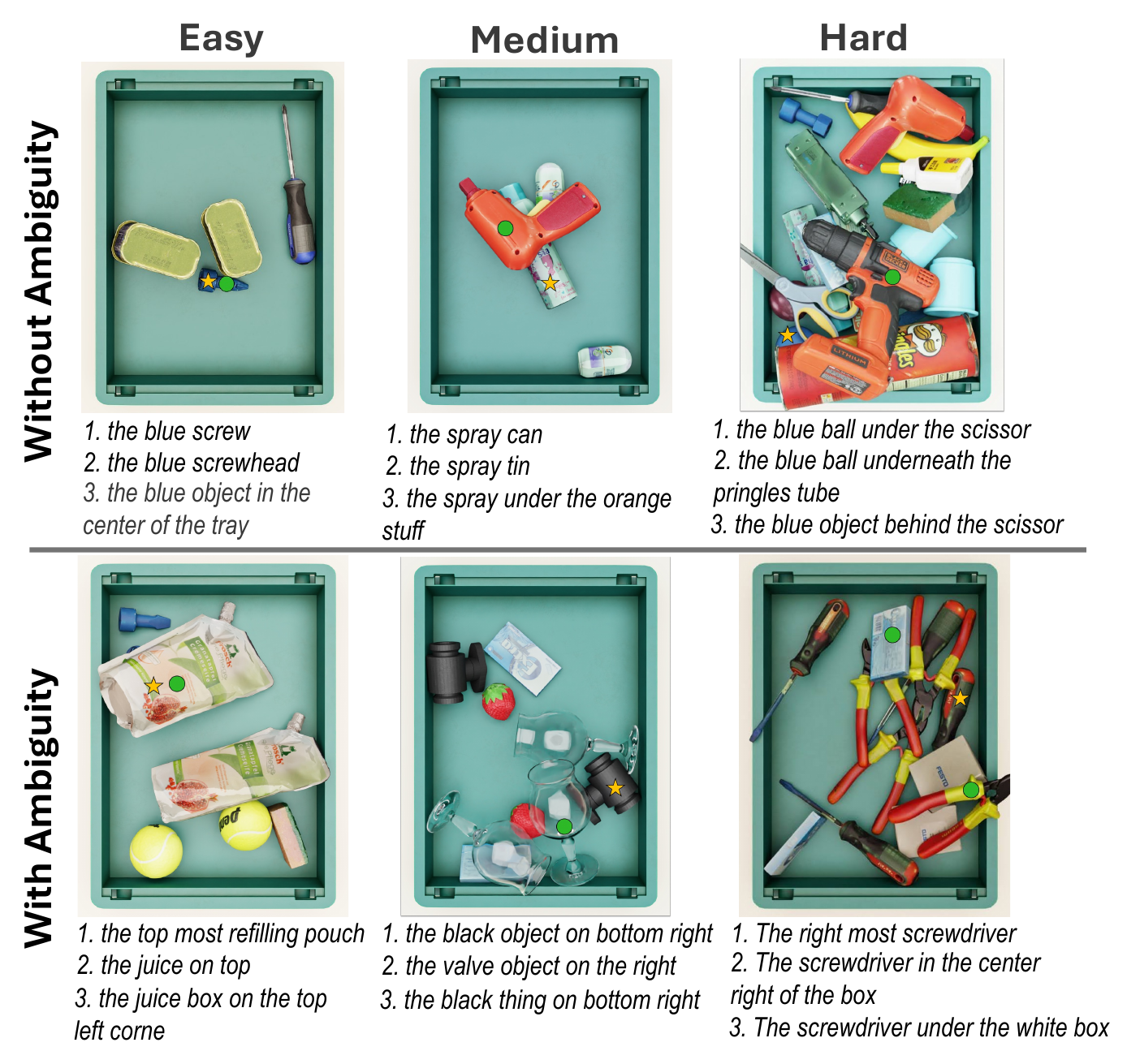}
    \vspace{-2mm}
    \caption{Examples of \ourdatasetshort at different task difficulties with three user-provided instructions. {\color{yellow}\FiveStar} indicates the target object, and {\tiny{\color{green}\CircleSolid}} indicates the ground-truth objects to pick.} 
    \label{fig:dataset}
    \vspace{-3mm}
\end{figure}

\begin{figure}[t]
    \centering
    \includegraphics[width=\linewidth]{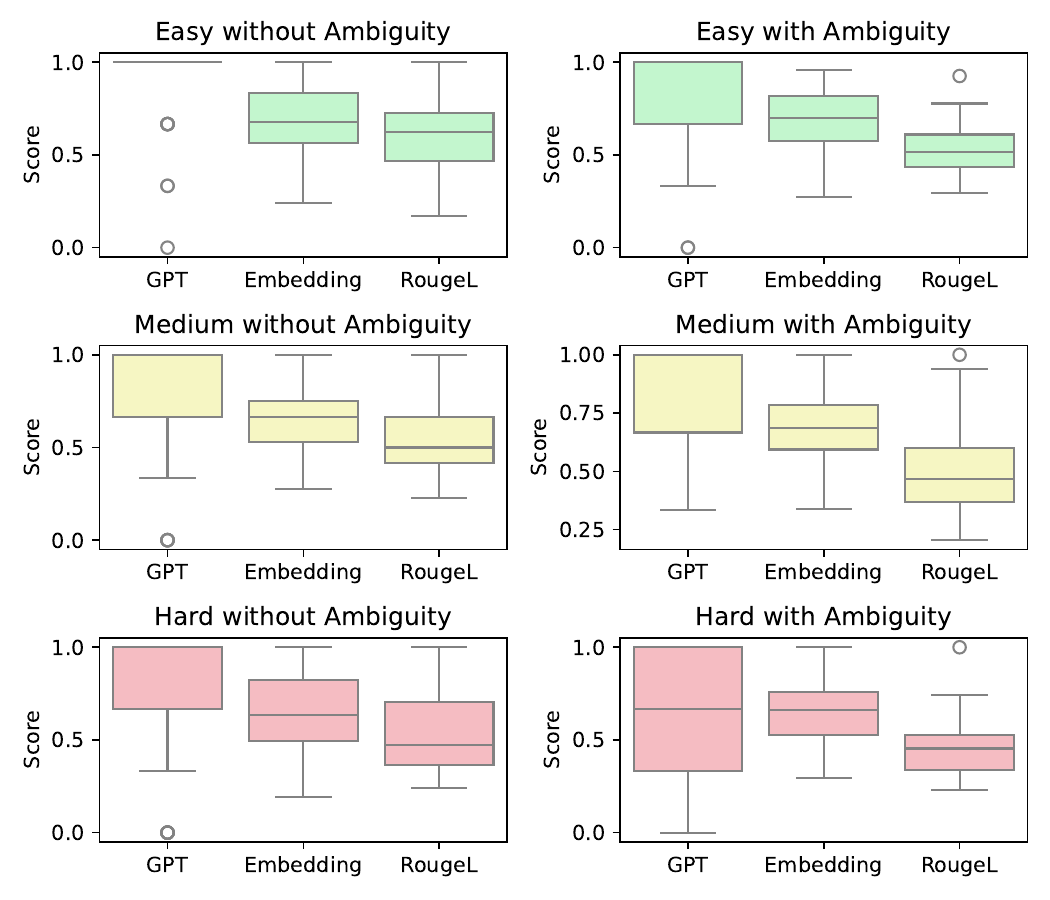}
    \vspace{-5mm}
    \caption{Similarity distribution among the three user-defined instructions used in the \ourdatasetshort scenarios.}
    \label{fig:dataset_ann_stats}
    \vspace{-5mm}
\end{figure}

We introduce the \ourdataset (\ourdatasetshort), a novel dataset built upon MetaGraspNetv2~\cite{MetaGraspNetV22024} to evaluate the robotic grasping task with free-form language instructions. 
MetaGraspNetv2 is a large-scale simulated dataset featuring challenging aspects of robot vision in the bin-picking setting, including multi-view RGB-D images and metadata, \eg object categories, amodal segmentation masks, and occlusion graphs indicating occlusion relationships between objects from each viewpoint. 
To build \ourdatasetshort, we selected scenes containing at least four objects to ensure sufficient scene clutter. 
\ourdatasetshort extends MetaGraspNetV2 in three aspects: i) we derive the ground-truth grasp sequence until reaching the target object from the occlusion graphs, ii) we categorize the task difficulty based on the obstruction level and instance ambiguity, and iii) we provide free-form language instructions collected from human annotators. 
Examples of occlusion graphs for real-world and FreeGraspData scenes are illustrated in \cref{fig:graph}.

%%%%%%%%%%%%%%%%%%%%%%%%%%%%%%%%%%%%%%%%%%%%%%%%%%%%%%%%%%%%%%
%%%%%%%%%%%%%%%%%%%%%%%%%%%%%%%%%%%%%%%%%%%%%%%%%%%%%%%%%%%%%%
\subsection{Ground-truth grasp sequence}
We obtain the ground-truth grasp sequence based on the object occlusion graphs provided in MetaGraspNetV2. 
As the visual occlusion does not necessarily indicate obstruction, we thus first prune the edges in the provided occlusion graph that are less likely to form obstruction. Following the heuristic that less occlusion indicates less chance of obstruction, we remove the edges where the percentage of the occlusion area of the occluded object is below $1\%$. 
From the node representing the target object, we can then traverse the pruned graph to locate the leaf node, which is the ground-truth object to grasp first. The sequence from the leaf node to the target node forms the correct sequence for the robotic grasping task.   

%%%%%%%%%%%%%%%%%%%%%%%%%%%%%%%%%%%%%%%%%%%%%%%%%%%%%%%%%%%%%%
%%%%%%%%%%%%%%%%%%%%%%%%%%%%%%%%%%%%%%%%%%%%%%%%%%%%%%%%%%%%%%

% \begin{figure*}[t]
% \centering
% \begin{overpic}[width=0.95\textwidth]{main/figures/occ.pdf}
%     \put(25,-1.5){\small (a) Real world}
%     \put(72,-1.5){\small (b) FreeGraspData}
% \end{overpic}
% \vspace{2mm}
% \caption{Occlusion graphs in representative scenarios.
% (a) Real-world scene and its corresponding occlusion graph.
% (b) Synthetic scene from FreeGraspData and its occlusion graph.
% Objects are colored by difficulty, and directed edges indicate occlusion dependencies.
% Example task instructions and step-by-step action plans are shown beneath each graph.}
% \label{fig:graph}
% \vspace{-4mm}
% \end{figure*}
%  (green: easy, yellow: medium, brown: hard)

\subsection{Grasp difficulty categorization}
We leverage the pruned occlusion graph to classify the grasping difficulty of target object into three levels:
\emph{``Easy"} refers to unobstructed target objects, \ie leaf nodes in the pruned occlusion graph; \emph{``Medium"} refers to objects obstructed by at least one object, \ie the maximum hop distance to the leaf nodes is 1; and
\emph{``Hard"} refers to objects obstructed by a chain of other objects, \ie the maximum hop distance to the leaf nodes is more than 1.
We label objects as \emph{``Ambiguous"} if multiple instances of the same class exist in the scene. 
Based on these criteria, we obtain six robotic grasping difficulty categories, also shown in \cref{fig:dataset}:
\emph{``Easy without Ambiguity"}, \emph{``Medium without Ambiguity"}, \emph{``Hard without Ambiguity"},
\emph{``Easy with Ambiguity"}, \emph{``Medium with Ambiguity"}, and \emph{``Hard with Ambiguity"}.

%%%%%%%%%%%%%%%%%%%%%%%%%%%%%%%%%%%%%%%%%%%%%%%%%%%%%%%%%%%%%%
%%%%%%%%%%%%%%%%%%%%%%%%%%%%%%%%%%%%%%%%%%%%%%%%%%%%%%%%%%%%%%
\subsection{Free-form language user instructions}
For each of the six difficulty categories, we randomly select 50 objects, resulting in 300 robotic grasping scenarios. 
For each scenario, we provide multiple users with a top-down image of the bin and a visual indicator highlighting the target object. 
% No additional context or information about the object is provided. 
% We instruct the user to provide an unambiguous natural language description of the indicated object with their best effort.
% In total, ten users are involved in the data collection procedure, with a wide age span. %and a balanced gender distribution. 
We randomly select three user instructions for each scenario, yielding a total of 900 evaluation scenarios.
This results in diverse language instructions, as shown in \cref{fig:dataset}.

\cref{fig:dataset_ann_stats} illustrates the similarity distribution among the three user-defined instructions in \ourdatasetshort, based on \gpt's interpretability, semantic similarity, and sentence structure similarity.
To assess \gpt's interpretability, we introduce a novel metric, the GPT score, which measures \gpt's coherence in responses.
For each target, we provide \gpt with an image containing overlaid object IDs and ask it to identify the object specified by each of the three instructions.
The GPT score quantifies the fraction of correctly identified instructions, ranging from 0 (no correct identifications) to 1 (all three correct).
We evaluate semantic similarity using the embedding score, defined as the average SBERT~\cite{reimers-2019-sentence-bert} similarity across all pairs of user-defined instructions.
We assess structural similarity using the Rouge-L score, computed as the average Rouge-L~\cite{lin2004automatic} score across all instruction pairs.
Results indicate that instructions referring to the same target vary significantly in sentence structure (low Rouge-L score), reflecting differences in word choice and composition, while showing moderate variation in semantics (medium embedding score).
Interestingly, despite these variations, the consistently high GPT scores across all task difficulty levels suggest that \gpt is robust in identifying the correct target in the image, regardless of differences in instruction phrasing.

\begin{table*}[t]
\centering
\renewcommand{\arraystretch}{0.9}
\tabcolsep 8.5pt
\caption{Experiments on \ourdatasetshort. Higher metric values (SSR and RSR) indicate better performance. Best performance under each setting is in \textit{italic}.}
\vspace{-2mm}
\label{tab:results_synt}
% \resizebox{\linewidth}{!}
{%
\begin{tabular}{lccccccccc}
\toprule
\multirow{2}{*}{Method} & \multirow{2}{*}{Reas.} & \multirow{2}{*}{Segm.}& \multirow{2}{*}{Metric} & \multicolumn{2}{c}{Easy} & \multicolumn{2}{c}{Medium} & \multicolumn{2}{c}{Hard} \\
\cmidrule(lr){5-6} \cmidrule(lr){7-8} \cmidrule(lr){9-10}
 & & & & w/o Amb. & w Amb. & w/o Amb. & w Amb. & w/o Amb. & w Amb. \\
\midrule
ThinkGrasp~\cite{qian2024thinkgrasp} &\checkmark & \checkmark &SSR & 0.63±0.02 & 0.46±0.02 & 0.13±0.03 & 0.16±0.02 & 0.05±0.02 & \textit{0.15±0.02} \\
\ourmethodshort  & \checkmark & \checkmark &SSR & \textit{0.64±0.03} & \textit{0.64±0.04} & \textit{0.40±0.04} & \textit{0.35±0.02} & \textit{0.13±0.01} & 0.13±0.02 \\
\midrule
\ourmethodshort & \checkmark(GT) & & RSR& \textit{0.83±0.02} & 0.77±0.02 &\textit{0.46±0.03} & 0.31±0.06 & 0.21±0.01 & \textit{0.16±0.04} \\
\ourmethodshort & \checkmark (Molmo) & & RSR & 0.83±0.06 & \textit{0.85±0.07} & 0.46±0.04 & \textit{0.33±0.04} & \textit{0.22±0.04} & 0.15±0.04 \\
\bottomrule
\end{tabular}
}
\vspace{-4mm}
\end{table*}

%%%%%%%%%%%%%%%%%%%%%%%%%%%%%%%%%%%%%%%%%%%%%%%%%%%%%%%%%%%%%
%%%%%%%%%%%%%%%%%%%%%%%%%%%%%%%%%%%%%%%%%%%%%%%%%%%%%%%%%%%%%
\section{Experiments}

We conduct experiments using both our synthetic evaluation dataset, \ourdatasetshort, and a gripper-equipped robotic arm in the real world.
We compare \ourmethodshort against the state-of-the-art method, ThinkGrasp~\cite{qian2024thinkgrasp}, which also uses \gpt for reasoning.
To comprehensively evaluate robotic grasping, we introduce novel metrics that assess performance based on both intermediate steps and the final robotic grasp.
Using \ourdatasetshort, we also perform extensive ablation studies on the key components of \ourmethodshort to validate our design choices.
Lastly, through real-world experiments, we demonstrate the effectiveness of \ourmethodshort in handling practical challenges such as imperfect depth measurements and robotic grasp execution. %More robotic demonstrations can be found in the Supplementary Material.

%%%%%%%%%%%%%%%%%%%%%%%%%%%%%%%%%%%%%%%%%%%%%%%%%%%%%%%%%%%%%
%%%%%%%%%%%%%%%%%%%%%%%%%%%%%%%%%%%%%%%%%%%%%%%%%%%%%%%%%%%%%
\subsection{Experiments on \ourdatasetshort}

%%%%%%%%%%%%%%%%%%%%%%%%%%%%%%%%%%%%%%%%%%%%%%%%%%%%%%%%%%%%%
\noindent\textit{Experimental setup.}  
We compare \ourmethodshort and ThinkGrasp on \ourdatasetshort, which includes 900 evaluation scenarios across the six task difficulty levels outlined in Sec.~\ref{sec:dataset}.
Both methods receive a top-down RGB image as input, along with user-defined task instructions.
Since each object in the evaluation set is associated with three different user instructions, we report the final performance metrics as the mean and standard deviation across these three instructions.

\vspace{1mm}
%%%%%%%%%%%%%%%%%%%%%%%%%%%%%%%%%%%%%%%%%%%%%%%%%%%%%%%%%%%%%
\noindent\textit{Performance metrics.} 
Since \ourdatasetshort is a static dataset, we mainly evaluate grasp reasoning and object segmentation.
We do not assess grasp estimation with \ourdatasetshort, as this aspect is best evaluated through robotic execution (Sec.~\ref{subsection:Exp_Real_World}).
For grasp reasoning, we report the Reasoning Success Rate (RSR), \ie the proportion of successful episodes in which the predicted object ID matches one of the ground-truth IDs.
For object segmentation, we report the Segmentation Success Rate (SSR), \ie the proportion of successful episodes in which the output mask achieves an Intersection over Union of at least 0.5 with the ground-truth mask.
Since ThinkGrasp does not produce object IDs, we can only report its SSR.
% For \ourmethodshort, we report both RSR and SSR. 
Moreover, we evaluate a variant of \ourmethodshort to assess the impact of object localization on reasoning, as measured by RSR.
This variant uses ground-truth (GT) localization instead of Molmo.

\vspace{1mm}
%%%%%%%%%%%%%%%%%%%%%%%%%%%%%%%%%%%%%%%%%%%%%%%%%%%%%%%%%%%%%
\noindent\textit{Discussion.} 
Tab.~\ref{tab:results_synt} presents the results for SSR and RSR using \ourdatasetshort.
\ourmethodshort significantly outperforms ThinkGrasp across almost all difficulty levels, except in ``\textit{Easy without Ambiguity}" scenario where both methods perform similarly.
This shows that \ourmethodshort effectively handles object ambiguity, thanks to its object localization design and mark-based visual prompting.
As scene clutter increases (Medium \& Hard scenarios), we observe that ThinkGrasp performs better in ambiguous cases than in non-ambiguous ones.
A qualitative analysis of the results suggests that this is due to ThinkGrasp’s tendency to segment all instances of the user-specified object.
In fact, the larger the number of ambiguities (\ie multiple instances of the target class), the higher the likelihood that the ground-truth obstructor belongs to the same class as the user-specified object.
Lastly, when comparing Molmo-based object localization with ground-truth (GT) detections, we find that Molmo achieves a higher RSR than GT localization.
This is because Molmo tends to miss highly occluded objects, increasing the likelihood that the reasoning module will select objects on the surface.

% \begin{figure}[t]
%     \centering
%     \includegraphics[width=0.45\textwidth]{main/figures/real-dataset.pdf}
%     \vspace{-2mm}
%     \caption{Samples from real-world experiments for different task difficulties. {\color{yellow}\FiveStar} indicates the user-described target object, and {\tiny{\color{green}\CircleSolid}} are the GT objects to pick.}
%     \label{fig:realworld}
%     \vspace{-5mm}
% \end{figure}

% \begin{figure*}[t]
% \centering
% \includegraphics[width=\textwidth]{main/figures/qualitative/qualitative_real_world_1.pdf}
% \includegraphics[width=\textwidth]{main/figures/qualitative/qualitative_real_world_2.pdf}
% \begin{overpic}[width=\textwidth, trim=0 -5cm 0cm 0, clip]{main/figures/qualitative/qualitative_synth_1.pdf}
%     \put(12,0){Input}
%     \put(22,0){Molmo detections}
%     \put(46,0){\gpt Reasoning}
%     \put(71,0){Segmentation}
%     \put(87,0){Estimated Pose}
% \end{overpic}
% \vspace{-2mm}
% \caption{Examples of three successful scenarios with \ourmethodshort:
% At the top, we show a real-world case where segmentation, pose, and motion were all correctly identified. In the middle, we present another real-world scenario where segmentation was incorrect, but GraspNet was still able to identify the correct pose to remove the obstacle. At the bottom, we display a successful example from a \ourdatasetshort scenario, featuring a highly cluttered environment.}
% \label{fig:qualitative} 
% \vspace{-4mm}
% \end{figure*}

%%%%%%%%%%%%%%%%%%%%%%%%%%%%%%%%%%%%%%%%%%%%%%%%%%%%%%%%%%%%%
%%%%%%%%%%%%%%%%%%%%%%%%%%%%%%%%%%%%%%%%%%%%%%%%%%%%%%%%%%%%%
\subsection{Experiments in the real world}\label{subsection:Exp_Real_World}

\begin{figure}[t]
    \centering
    \includegraphics[width=0.45\textwidth]{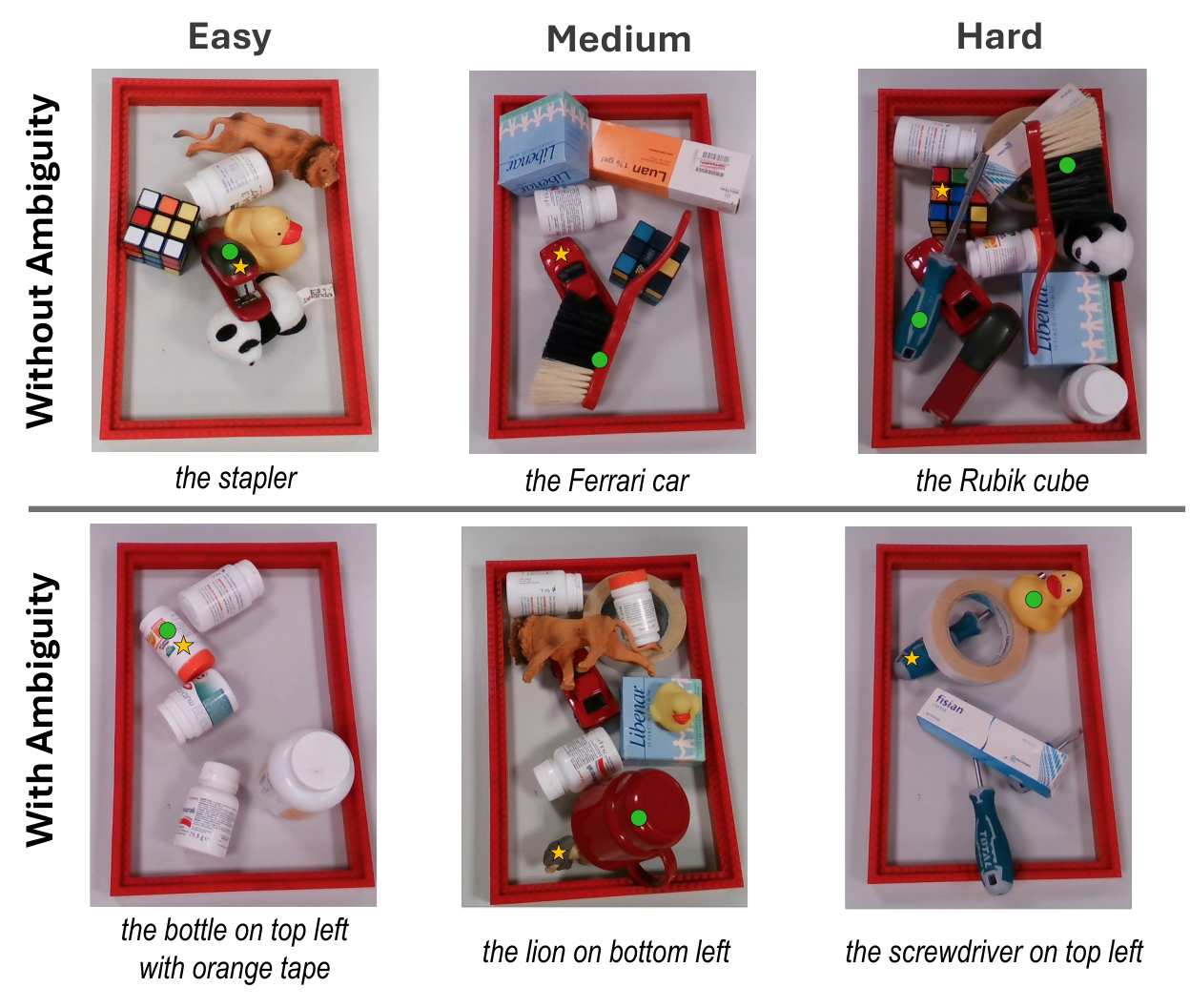}
    \vspace{-2mm}
    \caption{Samples from real-world experiments for different task difficulties. {\color{yellow}\FiveStar} indicates the user-described target object, and {\tiny{\color{green}\CircleSolid}} are the GT objects to pick.}
    \label{fig:realworld}
    \vspace{-5mm}
\end{figure}

%%%%%%%%%%%%%%%%%%%%%%%%%%%%%%%%%%%%%%%%%%%%%%%%%%%%%%%%%%%%%
\noindent\textit{Experimental setup.} 
We evaluate \ourmethodshort and ThinkGrasp~\cite{qian2024thinkgrasp} in a real-world scenario with a UR5e robotic arm equipped with an OnRobot RG2 parallel gripper. 
The scene setup is similar as in prior work ThinkGrasp~\cite{qian2024thinkgrasp} for fair comparison.
% Specifically, the robotic arm is in a table-top setting. 
The scene is captured by a RealSense D415 RGB-D camera with a top-down view featuring the cluttered red bin, at a resolution of 1280$\times$720.
We implement the motion planning and control via the MoveIt motion planning framework with ROS.
Adhering to the definition of six task difficulties in Sec.~\ref{sec:dataset}, we compose ten evaluation scenarios per difficulty level (\cref{fig:realworld}), each featuring a unique arrangement of objects
% (\eg screwdrivers, stationery items, boxes, toys, etc) with a single free-form language instruction.

\vspace{1mm}
%%%%%%%%%%%%%%%%%%%%%%%%%%%%%%%%%%%%%%%%%%%%%%%%%%%%%%%%%%%%%
\noindent\textit{Evaluation protocol.} 
At each evaluation episode, the robotic arm is tasked to grasp and place the target object from its cluttered bin (red) to the empty bin (blue), upon receiving the user instruction describing the target in free-form language.

\begin{figure*}[t]
\centering
\includegraphics[width=\textwidth]{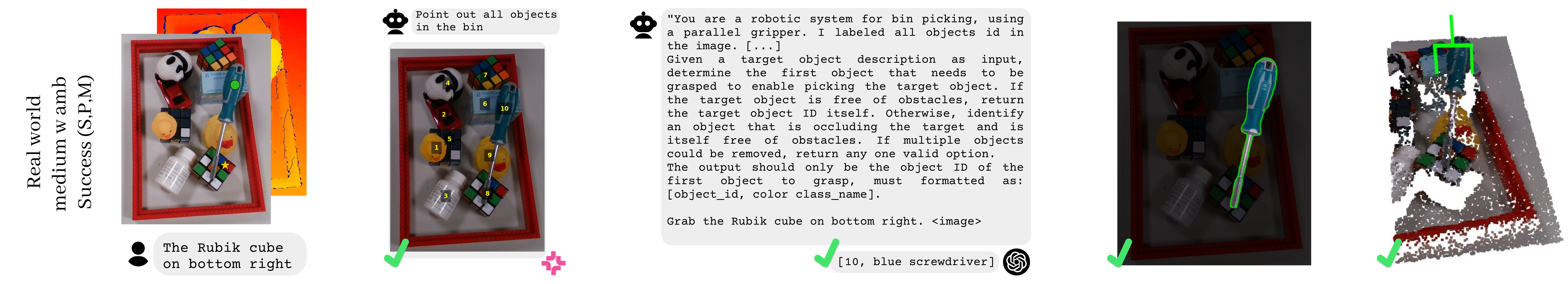}
\includegraphics[width=\textwidth]{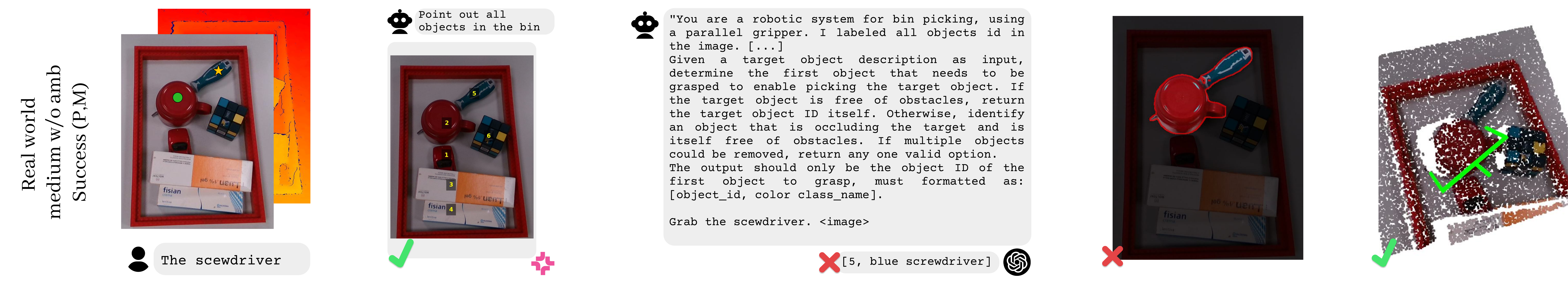}
\begin{overpic}[width=\textwidth, trim=0 -5cm 0cm 0, clip]{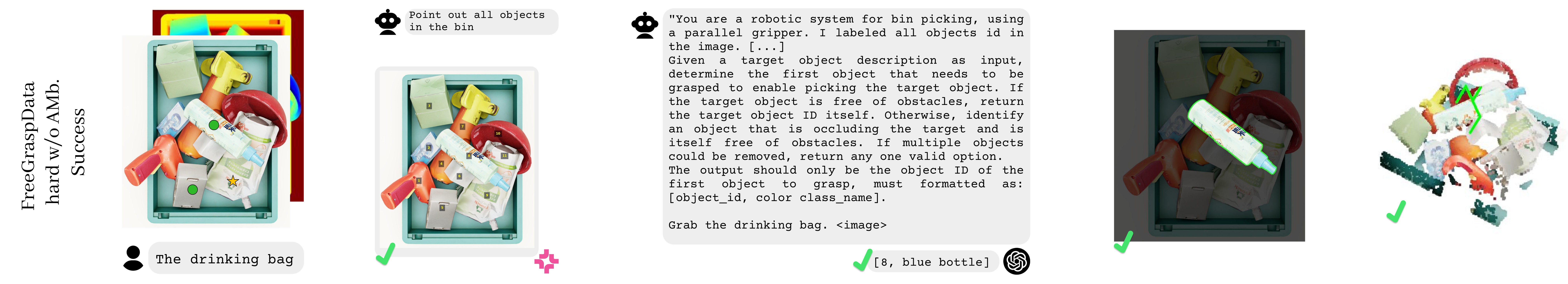}
    \put(12,0){Input}
    \put(22,0){Molmo detections}
    \put(46,0){\gpt Reasoning}
    \put(71,0){Segmentation}
    \put(87,0){Estimated Pose}
\end{overpic}
\vspace{-2mm}
\caption{Examples of three successful scenarios with \ourmethodshort:
At the top, we show a real-world case where segmentation, pose, and motion were all correctly identified. In the middle, we present another real-world scenario where segmentation was incorrect, but GraspNet was still able to identify the correct pose to remove the obstacle. At the bottom, we display a successful example from a \ourdatasetshort scenario, featuring a highly cluttered environment.}
\label{fig:qualitative} 
\vspace{-4mm}
\end{figure*}

The failure of the pipeline at each time step may occur at different modules, \ie vision-language grasp reasoning, grasp estimation, or robotic execution.
To comprehensively evaluate the pipeline's performance, we identify the failures at: 
i) Segmentation (S): when the predicted segmentation mask does not correspond to one of the GT unobstructed objects; 
ii) Pose (P): when the predicted pose does not correspond to one of the GT unobstructed objects; and 
iii) Motion (M): when robotic arm fails to retain one of the GT unobstructed objects in the gripper throughout the motion.
Interestingly, we find that a segmentation failure does not necessarily lead to a pose failure. 
As shown in \cref{fig:qualitative}, as long as the wrong segmentation mask includes a GT unobstructed object, the resulted pose can remain correct.

To study the impact of localized failures, we further devise three operational settings by defining the stopping criteria based on localized failures: 
i) \textit{Setting (S, P, M)} is the most stringent setting, where an evaluation episode will be terminated if any of the localized failures occur throughout the robotic reasoning and grasping task;
ii) \textit{Setting (P, M)} relaxes the segmentation failure, where an evaluation episode will be terminated only if a pose failure or motion failure occurs; and
iii) \textit{Setting (P)} relaxes the motion failure, where an evaluation episode will only be terminated if a pose failure occurs. 
% If motion failure under Setting~(P) occurs, we manually move the object to the bin.
% If no failure occurs, the episode continues until the target object is being grasped and placed.

\vspace{1mm}
%%%%%%%%%%%%%%%%%%%%%%%%%%%%%%%%%%%%%%%%%%%%%%%%%%%%%%%%%%%%%
\noindent\textit{Performance metrics.}
We use three metrics, inspired by navigation tasks~\cite{anderson2018evaluationAgents}, to measure the effectiveness and efficiency of ThinkGrasp and \ourmethodshort.
We define the \textit{success} of an episode when the correct target object is grasped and placed to its pre-defined destination.
The \textit{success rate (SR)} reports the ratio of successful episodes among all evaluation episodes, \ie $SR = \frac{N_s}{N}$, where $N$ is the number of total evaluation episodes and $N_s$ is the number of successful episodes. 
The \textit{path efficiency (SE)} reports the normalized inverse step counts, \ie $PE = \frac{1}{N_s} \sum_{i=1}^{N_s} \frac{l_i}{p_i}$, where $l_i$ is the number of minimum steps to solve the episode as defined by an oracle (\ie the operator), and $p_i$ is the actual number of steps the robot actually had to take to solve the episode.
Finally, the \textit{Success Weighted (normalized inverse) path length (SPL)}, calculated as $SPL = \frac{1}{N} \sum_{i=1}^{N} S_i \frac{l_i}{p_i}$, where $S_i$ is $1$ if episode $i$ is successful, $0$ otherwise.

\begin{table*}[t]
\renewcommand{\arraystretch}{0.9}
\centering
\caption{Results of real-world experiments. Higher metric values (SR, PE and SPL) indicate better performance. Best performance under each setting is in \textit{italic}.}
\label{tab:results_real}
\vspace{-2mm}
\newcolumntype{x}{>{\centering\arraybackslash}p{0.33cm}}
\newcolumntype{k}{>{\centering\arraybackslash}p{0.18cm}}
\begin{tabularx}{\linewidth}{X|kkk|xxx|xxx|xxx|xxx|xxx|xxx}
\toprule
\multicolumn{1}{c}{Method} & \multicolumn{3}{c}{} &\multicolumn{6}{c}{Easy} & \multicolumn{6}{c}{Medium} & \multicolumn{6}{c}{Hard}\\
\cmidrule(lr){5-10} \cmidrule(lr){11-16} \cmidrule(lr){17-22}
  \multicolumn{1}{c}{} & \multicolumn{3}{c}{Stop criteria} & \multicolumn{3}{c}{w/o Amb.} & \multicolumn{3}{c}{w Amb.}  & \multicolumn{3}{c}{w/o Amb.} & \multicolumn{3}{c}{w Amb.}  & \multicolumn{3}{c}{w/o Amb.} & \multicolumn{3}{c}{w Amb.}\\
  \cmidrule(lr){5-7} \cmidrule(lr){8-10} \cmidrule(lr){11-13} \cmidrule(lr){14-16} \cmidrule(lr){17-19} \cmidrule(lr){20-22}
 & S & P & M & SR & PE & SPL & SR & PE & SPL & SR & PE & SPL & SR & PE & SPL & SR & PE & SPL & SR & PE & SPL\\
 \midrule

ThinkGrasp~\cite{qian2024thinkgrasp} & \checkmark & \checkmark & \checkmark & \textit{0.60} & 1.0 & \textit{0.60} & 0.40 & 0.71 & 0.28 & 0.0 & 0.0 & 0.0 & 0.0 & 0.0 & 0.0 & 0.0 & 0.0 & 0.0 & 0.0 & 0.0 & 0.0\\
\ourmethodshort & \checkmark & \checkmark & \checkmark & 0.50 & 1.0 & 0.50 & \textit{0.80} & \textit{0.85} & \textit{0.68} & \textit{0.20} & \textit{1.0} & \textit{0.20} & \textit{0.20} & \textit{1.0} & \textit{0.20} & 0.0 & 0.0 & 0.0 & \textit{0.10} & \textit{1.0} & \textit{0.10}\\
\midrule
% \rowcolor{gray!20}
ThinkGrasp~\cite{qian2024thinkgrasp} & & \checkmark & \checkmark & 0.70 & 1.0 & 0.70 & 0.40 & 0.71 & 0.28 & 0.10 & 1.0 & 0.10 & 0.0 & 0.0 & 0.0 & 0.0 & 0.0 & 0.0 & 0.0 & 0.0 & 0.0\\
\ourmethodshort & & \checkmark & \checkmark & 0.70 & 1.0 & 0.70 & \textit{0.80} & \textit{0.85} & \textit{0.68} & \textit{0.20} & 1.0 & \textit{0.20} & \textit{0.20} & \textit{1.0} & \textit{0.20} & \textit{0.10} & \textit{1.0} & \textit{0.10} & \textit{0.10} & \textit{1.0} & \textit{0.10}\\
\midrule
ThinkGrasp~\cite{qian2024thinkgrasp} & & \checkmark & & \textit{1.0} & \textit{0.95} & \textit{0.95} & 0.70 & 0.74 & 0.52 & 0.40 & 0.92 & 0.37 & 0.10 & 0.67 & 0.07 & 0.10 & 1.0 & 0.10 & 0.0 & 0.0 & 0.0\\
\ourmethodshort & & \checkmark & & 0.90 & 0.94 & 0.85 & \textit{0.80} & \textit{0.85} & \textit{0.68} & \textit{0.60} & \textit{0.92} & \textit{0.55} & \textit{0.40} & \textit{0.90} & \textit{0.36} & \textit{0.20} & 1.0 & \textit{0.20} & \textit{0.10} & \textit{1.0} & \textit{0.10} \\
\bottomrule
% \rowcolor{gray!20}

\end{tabularx}
\vspace{-4mm}
\end{table*}

\vspace{1mm}
%%%%%%%%%%%%%%%%%%%%%%%%%%%%%%%%%%%%%%%%%%%%%%%%%%%%%%%%%%%%%
\noindent\textit{Discussion.}
Tab.~\ref{tab:results_real} reports the results of \ourmethodshort and ThinkGrasp, in real-world experiments across three operational settings.
\ourmethodshort outperforms ThinkGrasp in nearly all evaluated settings and task difficulties, except in the least difficult scenario, \ie \textit{Easy without Ambiguity}.
ThinkGrasp struggles to identify and segment the correct object to grasp as scene clutter and object ambiguity increase. 
Under the most stringent \textit{Setting (S, P, M)}, ThinkGrasp fails all evaluation episodes in the \textit{Medium} and \textit{Hard} scenarios.
In contrast, \ourmethodshort performs better in these challenging scenarios, thanks to the \textit{vision-language grasp reasoning} module, which helps identify the correct object to pick.

In \textit{Setting (P, M)}, we sometimes observe slightly better performance than in \textit{Setting (S, P, M)}. 
This occurs when the segmentation mask includes multiple objects (which counts as a segmentation failure), and one of them is a ground-truth obstructed object. 
In such cases, the estimated grasp pose is more likely to be correct, leading to more successful episodes.
We qualitatively demonstrate such a case in \cref{fig:qualitative}, where the second row shows a case with incorrect segmentation but a correct estimated pose. 
In contrast, the first row shows a case where both segmentation and estimated pose are correct.

Lastly, as expected, when robotic execution is not considered (\ie under \textit{Setting (P)}), we observe the best performance.
Note that, the lower performance in the \textit{Medium} and \textit{Hard} scenarios highlights the difficulty of this robotic reasoning and grasping task, which we deem far from being solved. 
Future work in visual spatial reasoning is necessary to better address challenges like scene clutter and ambiguity, which are critical in real-world applications.

%%%%%%%%%%%%%%%%%%%%%%%%%%%%%%%%%%%%%%%%%%%%%%%%%%%%%%%%%%%%%
%%%%%%%%%%%%%%%%%%%%%%%%%%%%%%%%%%%%%%%%%%%%%%%%%%%%%%%%%%%%%
\subsection{Computational analysis}
\ourmethodshort ran on a workstation equipped with a 24 GB NVIDIA RTX 4500 GPU.
The computational analysis of \ourmethodshort was conducted using 60 episodes, covering all task difficulties in our real-world experiments.
The mean total execution time is 15.39 seconds, with the breakdown of time for each main component as follows: \textit{object localization} with Molmo (9.12s), \textit{grasp reasoning} with \gpt (5.46s), \textit{object segmentation} with LangSAM (0.71s), and \textit{grasp estimation} with GraspNet (0.10s).
Since the camera is externally mounted, VLM-based reasoning and pose estimation can be performed in parallel with robotic manipulation after the first step of the episode, while the robot is in motion.

\section{Conclusions}

\noindent We introduced \ourmethodshort, a novel approach that leverages pre-trained \vlms for robotic grasping by interpreting free-form instructions and reasoning about spatial relationships. 
Our investigation showed that while \vlms, \eg \gpt, are known for strong general reasoning capabilities, they struggle with visual spatial reasoning, highlighting an important gap for both visual grounding and spatial awareness.
\ourmethodshort was designed to mitigate such gap with mark-based visual prompting and contextualized reasoning, outperforming the state-of-the-art, ThinkGrasp, in both synthetic and real-world robotic validations.
Overall, \ourmethodshort demonstrates the potential of combining \vlms with modular reasoning to tackle robotic grasping challenges in real-world scenarios.

\noindent\textit{Limitations and future works.}
A key limitation we observed of \ourmethodshort is GPT-4o’s limited visual-spatial capability, particularly in understanding object occlusion.
While we tested specialized spatial VLMs like SpaceLLaVA~\cite{SpatialVLM2024}, they underperformed compared to GPT-4o.
Another limitation is the lack of a mechanism to adapt the initial human instructions as scenes change with object removal.
% For example, prompts like \texttt{the duck on the right of the screwdriver} become invalid if the screwdriver is removed. 
Future work could explore memory mechanisms to track scene changes or adaptive instruction updates using VLMs to dynamically adjust references as objects are removed.

{
    \small
    \bibliographystyle{IEEEtranBST/IEEEtran}
    \bibliography{main}
}

\end{document}